\documentclass[letterpaper, 10 pt, conference]{ieeeconf}  

\IEEEoverridecommandlockouts                              

\usepackage{graphics} 
\usepackage{epsfig} 
\usepackage{mathptmx} 
\usepackage{times} 
\usepackage{amsmath} 
\usepackage{amssymb}  
\usepackage{txfonts}
\usepackage{algorithm} 
\usepackage{algpseudocode} 
\DeclareMathOperator*{\argmax}{\arg\!\max}

\title{\LARGE \bf
Feedback-Based Dynamic Feature Selection for Constrained Continuous Data Acquisition
}

\author{Alp Sahin and Xiangrui Zeng
\thanks{Authors are with Robotics Engineering Department,
        Worcester Polytechnic Institute, 100 Institute Road Worcester MA, United States
        {\tt\small \{asahin,xzeng2\}@wpi.edu}. Corresponding author: Xiangrui Zeng.}
}

\begin{document}

\maketitle
\thispagestyle{empty}
\pagestyle{empty}

\begin{abstract}
Relevant and high-quality data are critical to successful development of machine learning applications.
For machine learning applications on dynamic systems equipped with a large number of sensors, such as connected vehicles and robots, how to find relevant and high-quality data features in an efficient way is a challenging problem. 
In this work, we address the problem of feature selection in constrained continuous data acquisition. 
We propose a feedback-based dynamic feature selection algorithm that efficiently decides on the feature set for data collection from a dynamic system in a step-wise manner. 
We formulate the sequential feature selection procedure as a Markov Decision Process.
The machine learning model performance feedback with an exploration component is used as the reward function in an $\epsilon$-greedy action selection. 
Our evaluation shows that the proposed feedback-based feature selection algorithm has superior performance over constrained baseline methods and matching performance with unconstrained baseline methods.

\end{abstract}

\section{INTRODUCTION}

Machine learning has become a popular method for many challenging engineering problems such as predictive maintenance, diagnostics, and prognostics, etc.
The development of machine learning applications usually requires that a large amount of data is available or can be collected with some effort. 
Although big data is becoming available in many areas, there are still applications where the data collection is constrained by the cost of data storage, data transmission, and the limited bandwidth. 
In these cases, capturing relevant and high-quality data in an efficient manner is desired.

When machine learning methods are applied using existing datasets, most of the real-world challenges on data acquisition can be overlooked. 
In a real-world setting, constraints placed by the data collection equipment, storage, time and human resources limit the collection and labeling of the data. 
There is also the consideration of data flow, where the instances of data can only become sequentially available after a delay.
In these cases, it might be advantageous to consider a feedback-based approach, where the data collection decisions depend on the previous decisions and collected data.

\begin{figure}[t!]
    \centering
    \includegraphics[scale=0.17]{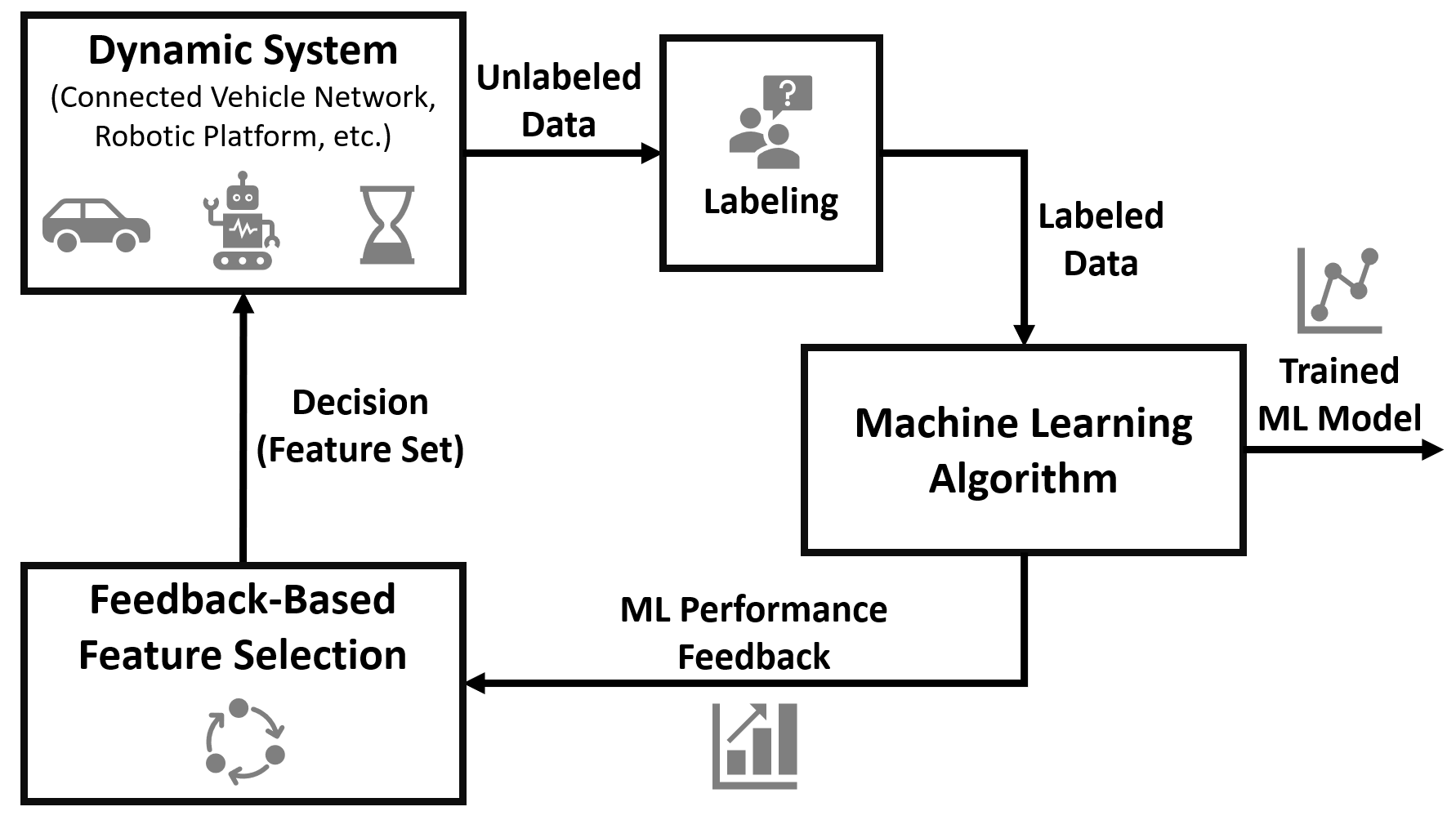}
    \caption{Process scheme for feature selection, data collection and machine learning framework.}
    \label{fig:outer_loop_draft}
\end{figure}

Specifically, we focus on the machine learning applications such as diagnostics and prognostics of connected vehicles and robots. 
Modern connected vehicles and robots may be equipped with many sensors and these sensors may generate as much as Gigabytes of data per minute.
Due to the on-board storage and computation limitations as well as data communication constraints, it is usually not possible to collect data from the entire set of sensors all the time. 
Therefore only a limited number of features can be collected from each vehicle or robot over a period.
The number of vehicles and robots are also limited, bounding the number of instances that can be collected. 
Furthermore, it usually takes weeks or months to collect the required data for diagnostics and prognostics. 
Therefore, it is important to make feature selection decisions based on the collected data and dynamically adjust the features to be collected in this on-going data acquisition process. 
The goal of this research is to design a feature selection policy that yields an improved machine learning performance, when data is acquired according to the policy. 
Fig. \ref{fig:outer_loop_draft} illustrates the described data acquisition problem.
Existing work in literature does not tackle this problem directly. 
We explain how certain aspects of the problem are partially tackled and solved in the detailed literature review in Section \ref{sec:RW}.

In this paper, a novel closed-loop feature selection and data acquisition strategy is proposed, where the new data collection decisions are made based on the feedback from collected data. 
This strategy includes a feature selection algorithm based on the wrapper method and extended by an exploration component. 
At each step, features are selected sequentially starting from an empty set such that they yield the highest reward in an $\epsilon$-greedy manner. 
Then new data in selected feature subset are collected.
To evaluate the performance of the developed approach, the proposed method is tested in a benchmarking simulation with several other baseline methods.

The rest of this paper is organized as follows. Section \ref{sec:RW} provides a brief review of the literature on techniques related to proposed problem. 
Section \ref{sec:Method} formally introduces the problem and formulates the feature selection solution. 
Section \ref{sec:BM} describes a list of baseline methods. 
Section \ref{sec:Eval} provides the evaluation methods. 
Section \ref{sec:Res} presents the results and the interpretation for the simulated data acquisition cases. 
Section \ref{sec:Conc} concludes the work.

\section{RELATED WORK}\label{sec:RW}

To the best of our knowledge, existing research only partially investigates the constrained data acquisition problem and tackles some of the feature selection challenges. 
In this section, we briefly review related work and explain how they are connected to our problem setting.

Research in active learning focuses on finding the most efficient techniques to select the queries to be labeled in a supervised learning problem.
A review of active learning literature is presented in \cite{Settles2009}. 
Some active learning research is propagated towards the direction of feature selection. 
In \cite{Liu2004} a selective sampling approach is developed to facilitate more efficient feature selection, when the dataset includes large number of instances. Authors use the filter-based feature selection algorithm ReliefF together with an active feature selection strategy based on \textit{kd}-tree structure. \cite{Raghavan2006} provides a different approach to active learning where the learner is able to query an oracle for feedback on both features and labels. They show that the learning performance is improved, when the oracle provides information on the most important features. Simultaneous feature selection and active learning approaches are investigated in \cite{Zhang2012} and \cite{Li2019}. Considering active learning and feature selection as simultaneous operations over the input matrix, they formulate and solve it as an optimization problem. Similarly, a dual problem is formulated and solved in \cite{Kong2011} for graph classification.

Design of experiments (DOE) is a statistical method that investigates the relationship between system parameters and its output. DOE aims to discover the relationship efficiently, using least number of experiments possible. When applied to feature selection problems, DOE can obviate the need for exhaustive search over the feature space and increase the efficiency of feature selection. In \cite{Kwak2002} and \cite{Yang2008}, Taguchi method is used to aid the search over the feature subsets. In \cite{Tang2018}, authors approached the feature selection problem from the feature interactions perspective and used DOE methods to find the most significant feature interactions. However, DOE based approaches only consider a fixed dataset setting, where the features and instances are available prior to the feature selection and learning.

Static feature selection is a widely investigated problem in machine learning. Review works in this field, provide an overview of the feature selection research from different perspectives. \cite{Chandrashekar2014} focuses on types of selection strategies such as filter, wrapper and embedded methods. In \cite{Cai2018} supervised unsupervised and semi-supervised feature selection methods are reviewed. \cite{Li2017_1} provides a review from data perspective, where they present several categories of feature selection problems depending on the type of the considered dataset. \cite{Li2017_2} reviews the different problem settings and presents some of the advanced problems in feature selection.

Although most of the work on feature selection considers a fixed dataset, approaches towards online feature selection are also developed. These approaches either consider a stream of data instances, a stream of features, or both. In \cite{Perkins2003}, a stream of features is considered while the number of instances is fixed. A stage-wise gradient descent method called grafting is developed. \cite{Wu2010} investigates feature redundancy and feature relevance and uses them as a measure to select features through a stream. In \cite{Hoi2012}, a stream of instances is considered, where the total number of features is constant. They further assume that all features are available for every instance. Truncation and sparse projection techniques are used to reduce the number of active features used in learning. In their following work \cite{Wang2014}, they introduce the partial input case, where only some of the features are available in a stream of instances. An $\epsilon$-greedy approach is used to select the features and collect the data. \cite{Yu2014} considers streaming features and develop a scalable feature selection for big data applications using the correlations between the features and filtering out the redundant ones. In \cite{Zhang2016}, a trapezoidal data stream is considered, where both the number of instances and features are free to increase. This is commonly the case in text classification tasks, where each additional document can also introduce new vocabulary. Authors combine the online learning and streaming feature selection approaches, where they carry out different updates using the existing features and new features. An imbalanced data case is investigated in \cite{Zhou2017}. They develop a method based on rough set theory to perform feature selection from a stream of features. In \cite{Gaudel2010} feature selection is formulated as a reinforcement learning problem and solved approximately using Monte Carlo Tree search and Upper Confidence Tree framework.

Some of the most recent work in feature selection includes budget-constrained and cost-sensitive problem settings. In \cite{Nushi2016} a budgeted data collection procedure is considered. Different from the active learning case, they consider the acquisition of features to have a cost and labeling procedure to be noisy. They develop a learning algorithm B-LEAFS, which keeps track of the uncertainty in the model parameters for each feature and runs a greedy decision-making process for budgeted learning and feature selection. A sequential and cost-sensitive feature acquisition problem is considered in \cite{Contardo2016},\cite{Shim2017} and \cite{Kachuee2019}. They formulate and solve the problem through reinforcement learning.

Related works presented in here mostly focus on reducing the dimensionality of the collected data instead of intercepting the machine learning procedure at the data acquisition step. 
Research in sequential feature acquisition does not account for the case where a process delay is present between the decision-making and data arrival. 
The problem considered in this work requires a decision-making on what features to acquire for training a machine learning model in the future, while accounting for the delay.
We believe that our work will present a novel and practical problem and an initial solution regarding real-world data acquisition for machine learning applications.

\section{METHODOLOGY}\label{sec:Method}

\subsection{Problem Statement}

\begin{figure}[t!]
    \centering
    \includegraphics[scale=0.17]{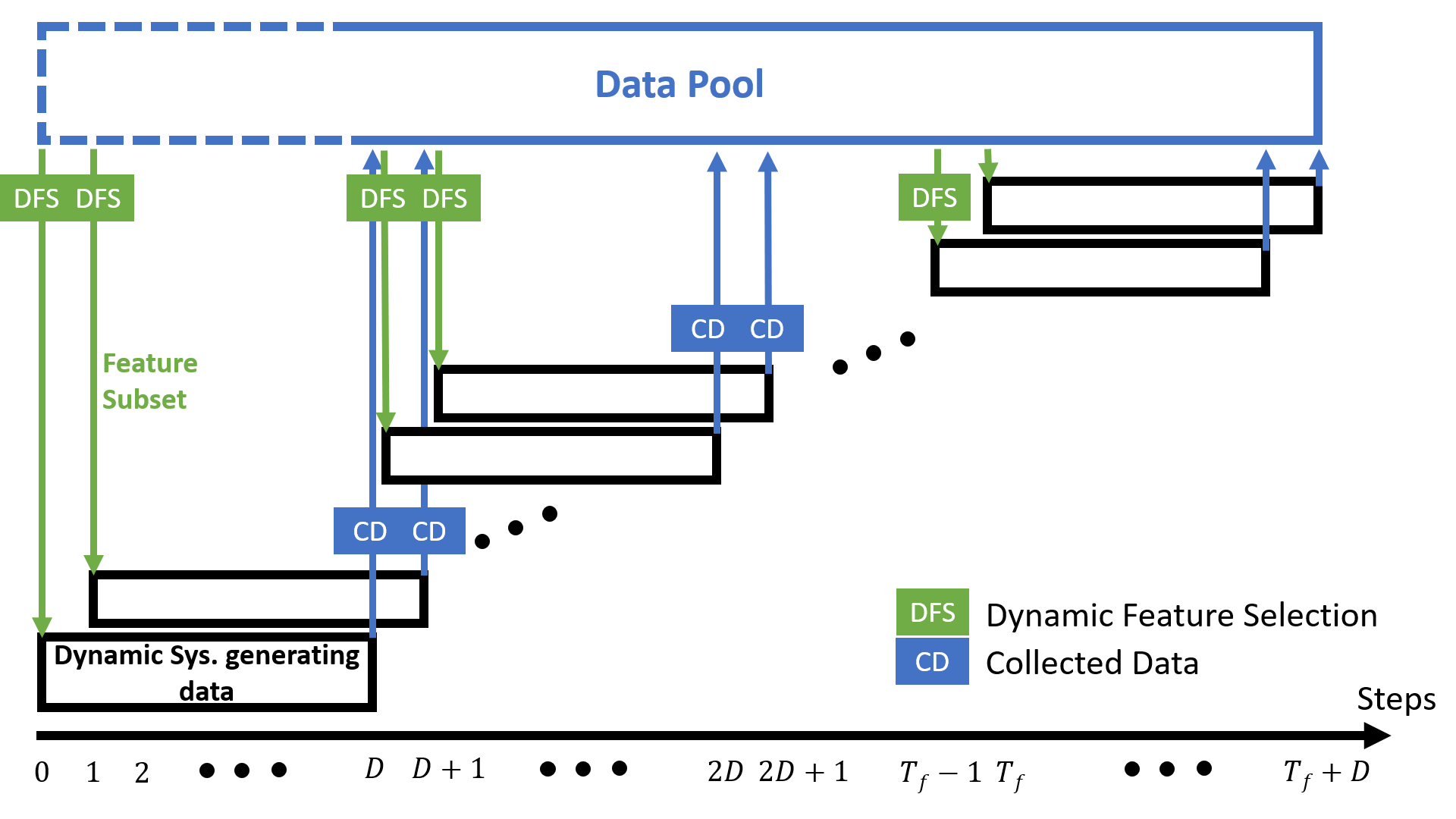}
    \caption{Delayed data collection process illustrated with respect to steps.}
    \label{fig:delayed collection}
\end{figure}
We consider a continuous data acquisition problem illustrated in Fig. \ref{fig:delayed collection}.
The goal is to design a feedback-based dynamic feature selection algorithm to decide what features to acquire in an on-going data collection process.
The input to the feedback-based dynamic feature selection algorithm is all the data that have been already collected, as well as the machine learning model performance based on these data.
The output of this algorithm is a feature set to be collected from the dynamic system.

There is a process delay, denoted by $D$, between the feature selection decision and the data acquisition.
The decision-making process is run for a total number of steps $T_f$. 
Let $S_t=\{i_1,i_2,...,i_k | i_j \in \mathbb{Z}^+, i_j \leq d \}$ be the feature selection decision at step $t$, where $d$ is the total number of collectable features, $k$ is the number of features to be collected, and $i_j$ denotes the index of a selected feature. 
Corresponding data $X_t \in \mathbb{R}^{n \times k}$ is collected at step $t+D$, where $n$ denotes the number of collected instances. 
After $X_t$ is collected, instances are labeled with binary labels, resulting in $y_t \in \mathbb{R}^n$.

Following assumptions are made:
\begin{itemize}
    \item The process delay $D$ and the total number of collectable features $d$ are known constants.
    \item The number of instances $n$ and the number of collected features $k$ depend on the system constraints and are constant for all $t$.
    \item As soon as the input matrix $X_t$ is collected, the label vector $y_t$ becomes available.
    \item There is no missing data in $X_t$ or $y_t$ for any $t$.
\end{itemize}
The goal of this work is to develop an algorithm for selecting $S_t$ at step $t$, depending on previous decisions $\mathcal{S}_t=\{S_1,S_2,...,S_{t-1}\}$, previously collected data $\mathcal{X}_t=\{X_1,X_2,...,X_{t-D-1}\}$, and corresponding labels $\mathcal{Y}_t=\{y_1,y_2,...,y_{t-D-1}\}$, such that a machine learning algorithm trained on the data collected before step $T_f+D$ will have adequate performance on classification tasks.

\subsection{Feature Selection Algorithm}

At one given time step $t$, we consider a multi-stage sequential feature selection problem, and formulate it as a Markov Decision Process (MDP). 
The states of this system are represented with $S_{t,j}$, where $j$ denotes the stage number in the feature selection sequence. 
Starting from state $S_{t,0}=\emptyset$, the feature selection algorithm sequentially approaches the termination state $S_{t,k}$, which is the feature selection decision to be used at step $t$. 
Note that $|S_{t,k}|=k$, where $k$ is the number of features to be selected.
The set of available actions depend on the previously selected actions. 
At the sequential feature selection stage $j$, the set of available actions is denoted by $A_{t,j}=\mathcal{F} \setminus S_{t,j-1}$, where $\mathcal{F}$ is the set of all features. 
If an action $a_j \in A_{t,j}$ is taken, the state is updated as $S_{t,j} = S_{t,j-1} \cup a_j$.

Our proposed solution to the feature selection problem with the given state-action space is analogous to the $\epsilon$-greedy algorithm in reinforcement learning. 
The reward function consists of the following two components.

\subsubsection{Machine Learning Performance}

The first component is a performance measure for the machine learning algorithm. In this work, we use the f1-score. 
However, any metric that yields a real-valued scalar is applicable. Resulting reward component is $r_m$, which is a function of the current state $S_{t,j-1}$, action $a_j$, and data in storage $\mathcal{X}_t$, $\mathcal{Y}_t$.

Let $S'$ denote the potential set of features for state-action pair $(S_{t,j-1},a_j)$, $S'=S_{t,j-1}\cup a_j$. 
Before training and testing the machine learning algorithm, relevant data needs to be extracted from $\mathcal{X}_t$, with feature set $S'$. 
For every $S_i$ in $\mathcal{S}_t$, it is checked whether $S' \subseteq S_i$. 
If $S'$ is a subset of some $S_i$, the data of $S_i$ can be used to estimate the machine learning performance using the feature set $S'$.
Available data for the evaluation of $S'$ is denoted with $X_{S'}$, 
$X_{S'} \in \mathbb{R}^{pn \times j}$, where $p$ is the number of times the data is collected using all the features in $S'$. 
The output vector $y_{S'}$ is obtained from $\mathcal{Y}_t$ accordingly. 
Available data is split into training and testing sets with a ratio of 0.8:0.2. 
f1-score is computed as $r_m$ using the results on the test set.
If $S'$ is not a subset of any $S_i \in \mathcal{S}_t$, there has been no data collected using the feature subset $S'$.
In this case, we can set $|X_{S'}|=0$ and $r_m=0$.

\subsubsection{A Term to Encourage Exploration}

To encourage exploration of features that do not have a large amount of samples, we include a second component $r_e$ in the reward function. 
The formulation is analogous to the Upper Confidence Bound (UCB) \cite{Sutton1998}, as shown in  (\ref{eq:exploration_measure}).
\begin{equation}
    r_e(t,N) = c \sqrt{\frac{\log(t)}{N+1}},
    \label{eq:exploration_measure}
\end{equation}
where $N$ denotes the number of instances that have been collected or in the progress of being collected with the newly selected feature $a$, and $c$ is a weight used to tune the trade-off between the exploration and performance components.

The resulting reward function is:
\begin{equation}
    r = r_m(S_{t,j-1},a_j,\mathcal{X}_{t-D},\mathcal{Y}_{t-D}) + r_e(t,a_j,\mathcal{S}_t).
\end{equation}

Optimal action ($a^*$) is defined as follows:
\begin{equation}
\begin{aligned}
    a^* &= \argmax_{a \in A_{t,j}} r\\ &= \argmax_{a \in A_{t,j}}[ r_m(S_{t,j-1},a,\mathcal{X}_{t-D},\mathcal{Y}_{t-D}) + r_e(t,a,\mathcal{S}_t)].
\end{aligned}
\end{equation}

According to the $\epsilon$-greedy selection, the optimal action is taken with a probability of $1-\epsilon$ and a random feature is selected from $A_j$ with a probability of $\epsilon$.

Provided pseudocode in Algorithm 1 summarizes the feature selection algorithm.
\begin{algorithm}[t!]
	\caption{Feature Selection at Time Step $t$} 
	\begin{algorithmic}[1]
		\State $S_{t,0} = \emptyset$
		\State $A_{t,0} = \mathcal{F}$
		\State initialize $r,r_m,r_e$
		\For {stage $j=1,2,\ldots,k$}
		    \State $A_{t,j} = A_{t,j-1} \setminus S_{t,j-1}$ 
		    \For{$a \in A_j$}
		        \State $S' = S_{t,j-1} \cup a$
		        \State get available data using $S'$
		        \If {no data}
		            \State $r_m(a)=0$
		        \Else
		            \State split train-test sets
		            \State train ML algorithm using training set
		            \State make predictions on test set
		            \State compute $r_m(a)$, using performance metric
		        \EndIf
		        \State compute $r_e(a)$
		        \State $r(a) = r_m(a) + r_e(a)$
            \EndFor
            \State $a^* = \argmax_{a} r(a)$
            \State with $p=1-\epsilon$, select optimally ($a^*$), $a_j = a^*$
            \State with $p=\epsilon$, select randomly ($a'$) from $A_{t,j}$, $a_j = a'$
            \State $S_{t,j} = S_{t,j-1} \cup a_j$
            \State reset reward vectors $r,r_m,r_e$
        \EndFor
        \State return $S_{t,k}$
	\end{algorithmic} 
\end{algorithm}
\subsection{Best Feature Subset Selection}\label{sec:BSS}
The criteria for the best feature subset selection at step $T_f+D$ is not the reward function in the sequential feature selection.
As a terminal feature selection decision needs to be made for the machine learning model at step $T_f$, the features with more data are favored over the features that need further exploration. 
For any $S$ in $\mathcal{S}_{T_f+D}$, available data $X_{S}$ can be extracted. 
In this case, we use cross-validation to compute the f1-score for each feature subset. 
Scores are weighted according to the number of collected instances $|X_S|$. 
Feature subset yielding the maximum weighted score is selected as the terminal decision, $S^*$, which is known as the best performing subset according to all the collected data.


\section{BASELINE METHODS} \label{sec:BM}
For a comparative analysis of our method, we established several baseline methods. 
Baseline methods are divided into two categories: the constrained methods and the unconstrained methods. 
Constrained baseline methods can be applied to the proposed problem without violating the constraints, whereas unconstrained baselines can only be applied after relaxing some of the constraints.

\subsection{Constrained methods}
\begin{itemize}
    \item C1: At each step $t$, the decision $S_t$ is generated randomly.
    \item C2: The feedback-based feature selection algorithm is used. However, to make a decision $S_{t+1}$, this baseline method waits for $X_t$ to arrive during the delay $D$. As the total time step $T_f$ stays unchanged, there are less iterations of data collection. Meanwhile, we increase the number of instances per step, $n_{C2}=\frac{n T_f (D+1)}{T_f + D}$, such that the size of the data has the same order of magnitude and is collected in same amount of time $T_f$.
    \item C3: It is assumed that the most important $k$ features are known prior to the data collection, either through domain knowledge or analysis of a previously collected data. Data is collected on the same features at each step. In this work, we determine the most important features by running a Random Forest classifier on some pre-collected data.
\end{itemize}

\subsection{Unconstrained methods}
\begin{itemize}
    \item UC1: No feature selection is used. Data from all $d$ features is collected at every step. The number of collected instances per step is $n_{UC1} = \frac{nk}{d}$, such that the size of the resulting data in storage $\mathcal{X}_t$ has the same order of magnitude with the data collected by feedback-based method.
    \item UC2: This baseline is similar to UC1. However, $n_{UC2} = n$, resulting in a larger dataset.
    \item UC3: Feedback-based feature selection algorithm is used. Similarly to C2, to make a decision $S_{t+1}$, this baseline method waits for $X_t$ to arrive. Data acquisition procedure continues until same amount of data is collected. The required time to collect the data will exceed $T_f$ and it is scaled by the amount of process delay $D$.
\end{itemize}

Our proposed method is expected to have a superior performance than the constrained baselines, and is designed to match the performance of the unconstrained methods.

\section{EVALUATION}\label{sec:Eval}

\subsection{Cross Validation on Collected Data}
After data collection is completed at $T_f$, we can determine the best performing subset using the procedure described in Section \ref{sec:BSS}. 
With the best feature subset $S^*$, relevant data $X_{S^*}$ is extracted.
Then we perform a cross validation on $X_{S^*}$ to measure the machine learning performance.
\subsection{Testing Future Performance on Test Data}
In our simulation evaluation, we can also use a preserved test data that are not included in the data collecting process described in Fig. \ref{fig:delayed collection}.
We train the machine learning model using $X_{S^*}$, and perform predictions on the test set.
This evaluation method requires additional data to be collected and labeled, which may or may not be applicable in a real-world scenario.
But it can be used in the simulation platform to test the performance of the proposed feedback-based dynamic feature selection method.

\section{RESULTS AND DISCUSSION}\label{sec:Res}
The proposed data collection strategy is meant to suggest an improvement in the most limited cases, where there is a large amount of process delay, small number of available instances per step and small number of features to be collected. 
These parameters are application dependent and can assume a wide range of values. 
We run the feedback-based data collection algorithm on several parameter sets to demonstrate the relationship between the machine learning performance and the system parameters. 
A Random Forest classifier is used as the machine learning method. 
We perform the cross-validation and future performance testing on the outputs from the simulation.

To test the performance of the proposed data acquisition strategy, we use a benchmarking framework which is designed to simulate continuous data collection with delay. 
Framework partitions a given fixed dataset according to the constraints of the data collection problem, and delivers the requested number of data instances with requested features after given amount of delay.
The simulation platform is supplied with the Adult dataset, from UCI Machine Learning Repository \cite{Dua2019}. The dataset contains around 50000 samples with 14 features each. Categorical features are transformed within the feature selection algorithm through one-hot encoding only when a categorical feature is selected.

In Table \ref{tab:table delay}, results are provided for varying delay. We use f1-score as a metric and report mean and standard deviation of 20 simulation runs on the same parameter set. CV denotes the cross-validation results, whereas Test denotes the results for the future performance tests. As expected, performance degrades slightly as the amount of delay increases. In Table \ref{tab:table instance}, the effect of number of instances on the performance is shown. Mean score increases and the variance decreases as the number of collected samples increase. In Table \ref{tab:table feature}, two cases are compared, $k=3$ and $k=5$. Naturally, having access to more features yields improved performance. In Table \ref{tab:table step}, the effect of the number of steps is demonstrated. Although there is no clear trend, performance is improved as the algorithm is run for more steps and more data is gathered.

\begin{table}[t!]
    \caption{Performance comparison for different amount of delay $D$.}
    \centering
    \begin{tabular}{|c|c|c|c|c|c|c|c|}
         \hline
         $T_f$ & $D$ & $n$ & $k$ & CV mean & CV std & Test mean & Test std  \\
         \hline
         100 & 10 & 10 & 3 & 0.503 & 0.118 & 0.459 & 0.097 \\
         100 & 25 & 10 & 3 & 0.518 & 0.140 & 0.491 & 0.096 \\
         100 & 50 & 10 & 3 & 0.454 & 0.158 & 0.403 & 0.138 \\ 
         100 & 75 & 10 & 3 & 0.455 & 0.195 & 0.405 & 0.149 \\ 
         \hline
    \end{tabular}
    \label{tab:table delay}
\end{table}

\begin{table}[t!]
    \caption{Performance comparison for different number of instances per step $n$.}
    \centering
    \begin{tabular}{|c|c|c|c|c|c|c|c|}
         \hline
         $T_f$ & $D$ & $n$ & $k$ & CV mean & CV std & Test mean & Test std  \\
         \hline
         100 & 75 & 10 & 3 & 0.455 & 0.195 & 0.402 & 0.149 \\ 
         100 & 75 & 25 & 3 & 0.494 & 0.124 & 0.500 & 0.102 \\
         100 & 75 & 50 & 3 & 0.477 & 0.128 & 0.487 & 0.135 \\ 
         100 & 75 & 100 & 3 & 0.514 & 0.083 & 0.520 & 0.068 \\ 
         100 & 75 & 200 & 3 & 0.530 & 0.068 & 0.534 & 0.071 \\ 
         \hline
    \end{tabular}
    \label{tab:table instance}
\end{table}

\begin{table}[t!]
    \caption{Performance comparison for different number of selected features $k$.}
    \centering
    \begin{tabular}{|c|c|c|c|c|c|c|c|}
         \hline
         $T_f$ & $D$ & $n$ & $k$ & CV mean & CV std & Test mean & Test std  \\
         \hline
         100 & 75 & 10 & 3 & 0.455 & 0.195 & 0.402 & 0.149 \\ 
         100 & 75 & 10 & 5 & 0.486 & 0.159 & 0.415 & 0.110 \\
         \hline
    \end{tabular}
    \label{tab:table feature}
\end{table}

\begin{table}[t!]
    \caption{Performance comparison for different number of steps $T_f$.}
    \centering
    \begin{tabular}{|c|c|c|c|c|c|c|c|}
         \hline
         $T_f$ & $D$ & $n$ & $k$ & CV mean & CV std & Test mean & Test std  \\
         \hline
         100 & 75 & 10 & 3 & 0.455 & 0.195 & 0.402 & 0.149 \\
         200 & 150 & 10 & 3 & 0.479 & 0.163 & 0.444 & 0.150 \\
         400 & 300 & 10 & 3 & 0.528 & 0.124 & 0.537 & 0.099 \\ 
         600 & 450 & 10 & 3 & 0.512 & 0.115 & 0.498 & 0.117 \\ 
         \hline
    \end{tabular}
    \label{tab:table step}
\end{table}

\begin{figure}[t!]
    \centering
    \includegraphics[scale=0.17]{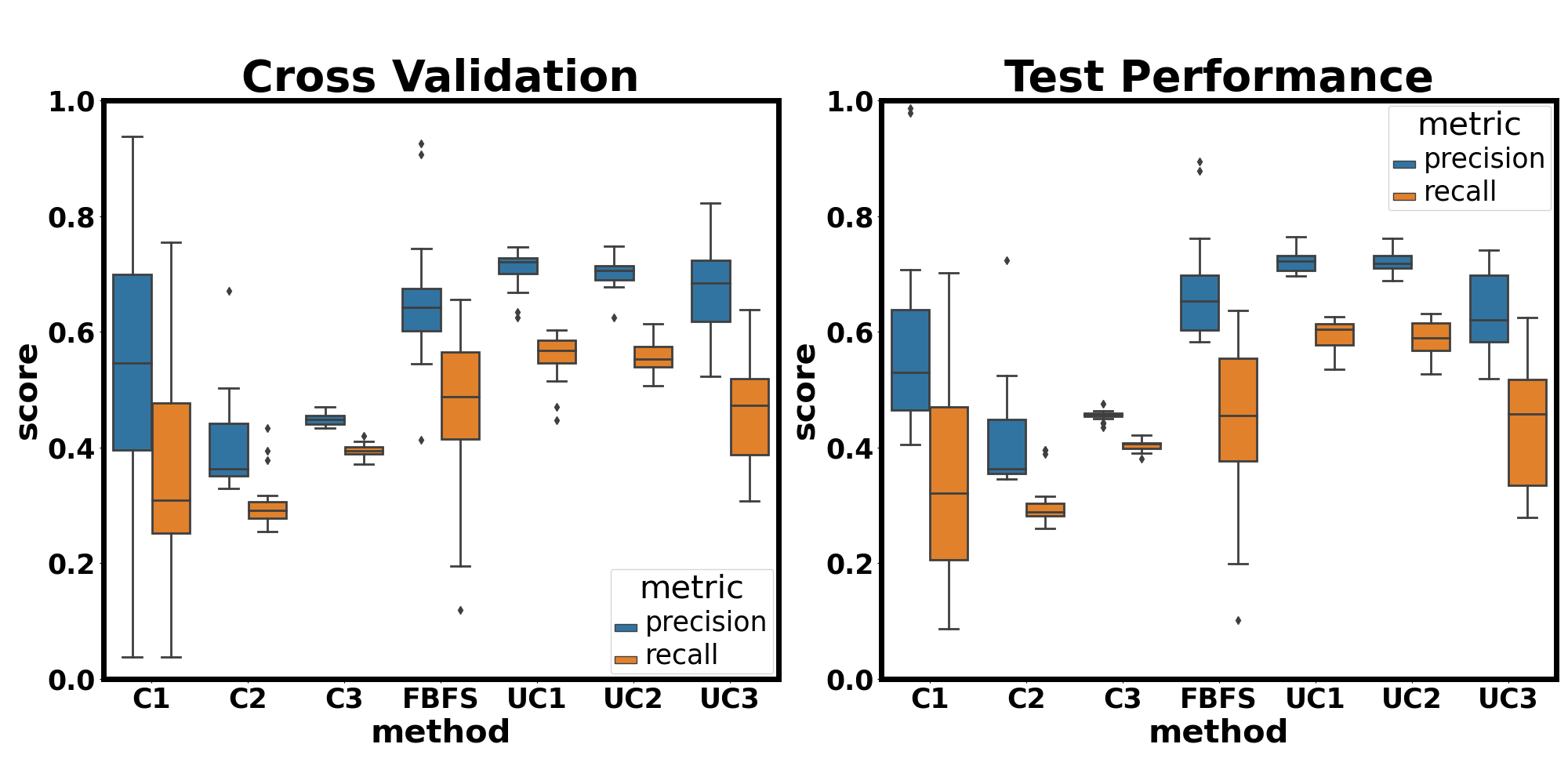}
    \caption{Comparison of precision and recall scores across proposed method and baseline methods for cross-validation and future performance test.}
    \label{fig:boxplot}
\end{figure}

For a selected set of parameters, $T_f = 600$, $D = 450$, $n = 10$ and $k = 3$, we run the simulation using Random Forest classifier for proposed baseline methods (C, UC) and the feedback-based method (FBFS). For each method, precision and recall scores in cross-validation and future performance testing are computed for 20 simulations. Distribution of the scores are provided in Fig. \ref{fig:boxplot} with boxplots.

\begin{figure}[t!]
    \centering
    \includegraphics[scale=0.15]{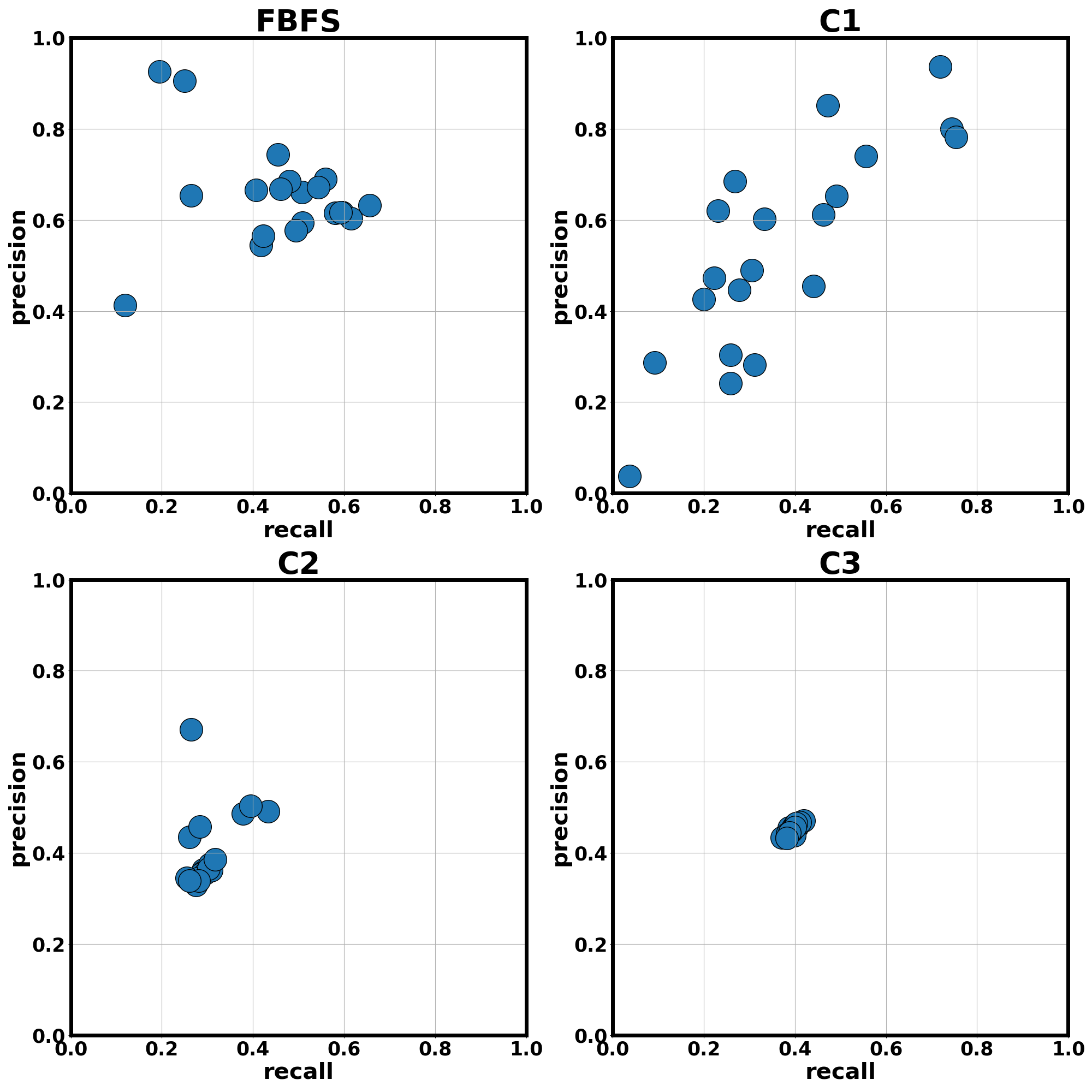}
    \caption{Comparison of precision vs recall across proposed method and constrained baseline methods for cross-validation.}
    \label{fig:vsC}
\end{figure}

\begin{figure}[t!]
    \centering
    \includegraphics[scale=0.15]{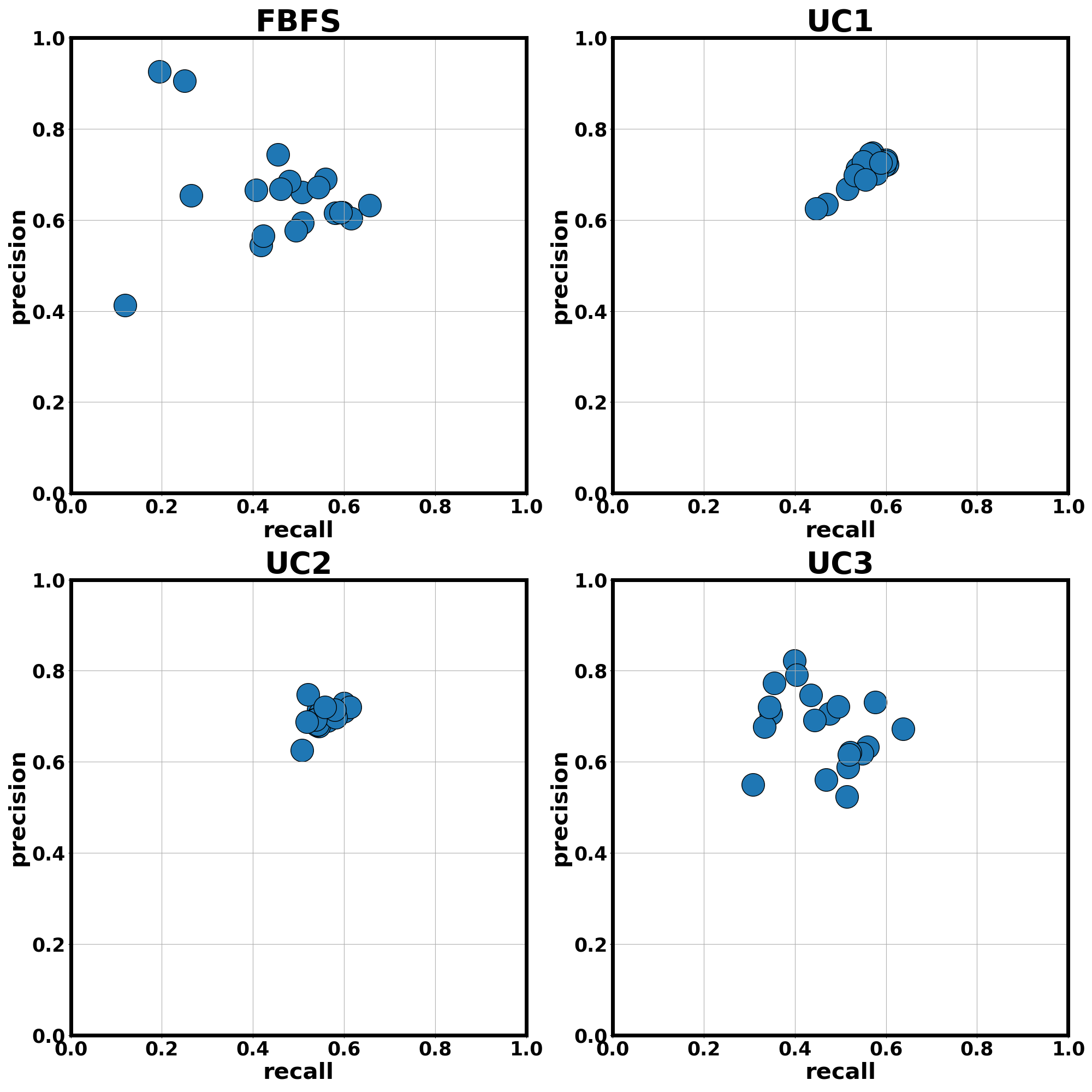}
    \caption{Comparison of precision vs recall across proposed method and constrained baseline methods for cross-validation.}
    \label{fig:vsUC}
\end{figure}

To show the combined precision and recall scores for each individual run, 2D scatter plots are used in Fig. \ref{fig:vsC} and Fig. \ref{fig:vsUC}. We only provide the cross-validation results in this case, since future test performances are similar as shown in Fig. \ref{fig:boxplot}.

It can be seen from Fig. \ref{fig:boxplot} that the 25th percentile performance of the proposed method in terms of machine learning model precision and recall beats the median performance of all constrained baselines. 
The median performance of the proposed method is close to that of the unconstrained baselines.
Fig. \ref{fig:vsC}, and Fig. \ref{fig:vsUC} show that, although with a relatively large variance, the proposed feedback-based method generally performs better than constrained baseline methods and show similar performance with the unconstrained methods in terms of precision and recall.




\section{CONCLUSIONS}\label{sec:Conc}
In this work, we proposed a constrained and continuous data acquisition problem for machine learning applications, that accounts for the real-world limitations in equipment, computation, storage and time. 
The developed solution aims to improve the machine learning performance through a feedback-based dynamic feature selection algorithm based on MDP formulation and $\epsilon$-greedy action selection. 
Our method is compared to constrained and unconstrained baseline methods and simulations showed that feedback-based method performs better than the constrained baselines and similarly with the unconstrained methods.











\end{document}